\documentclass[letterpaper, 10 pt, conference]{ieeeconf}
\IEEEoverridecommandlockouts

\usepackage{cite}
\usepackage{amsmath,amssymb,amsfonts}
\usepackage{algorithmic}
\usepackage{graphicx}
\usepackage{textcomp}
\usepackage{censor}
\usepackage{xcolor}
\usepackage{booktabs}   
\usepackage{array}
\usepackage{tabularx}   
\usepackage{bm}
\usepackage{tikz}
\usetikzlibrary{automata, positioning, arrows.meta}
\usepackage[format=hang, margin=10pt, font=small]{subcaption}

\definecolor{blockblue}{RGB}{100,149,237}
\definecolor{blockgray}{RGB}{220,220,220}
\definecolor{blockorange}{RGB}{255,179,71}  
\definecolor{blockgray}{RGB}{220,220,220}

\def\BibTeX{{\rm B\kern-.05em{\sc i\kern-.025em b}\kern-.08em
    T\kern-.1667em\lower.7ex\hbox{E}\kern-.125emX}}

\makeatletter
\newcommand{\linebreakand}{%
\end{@IEEEauthorhalign}%
\hfill\mbox{}\par
\mbox{}\hfill\begin{@IEEEauthorhalign}%
}
\makeatother
\newcommand\copyrighttext{%
\footnotesize \copyright 2026 IEEE. Personal use of this material is permitted. Permission from IEEE must be obtained for all other uses, in any current or future media, including reprinting/republishing this material for advertising or promotional purposes, creating new collective works, for resale or
redistribution to servers or lists, or reuse of any copyrighted component of this work in other works.}
\newcommand\copyrightnotice{%
\begin{tikzpicture}[remember picture,overlay]
\node[anchor=north,yshift=-1cm] at (current page.north) {\parbox{\dimexpr\textwidth-\fboxsep-\fboxrule\relax}{\centering \copyrighttext}};
\end{tikzpicture}%
}
\begin{document}

\title{\LARGE \bf
Driving Context-guided Model Predictive Planning and Control for Autonomous Car Racing at the Limit and Beyond
}

\author{ Ayoub Raji$^{1}$, Federico Sacco$^{2}$, Nicola Musiu$^{1}$, Marko Bertogna$^{1}$
\thanks{$^{1}$Ayoub Raji, Nicola Musiu and Marko Bertogna are with University of Modena and Reggio Emilia, 41125 Modena, Italy. {\tt\small ayoub.raji@unimore.it}}%
\thanks{$^{2}$Federico Sacco is with Hipert S.r.l.,
        Modena, Italy.}%
}

\maketitle
\copyrightnotice

\begin{abstract}

This paper presents a Model Predictive Control-based motion planning and control pipeline for autonomous car racing capable of adapting to different driving contexts, such as overtaking, nominal driving, and countersteering. A Cost Blending state machine manages the identification of different driving contexts and the selection of their predefined weights to be applied to the Model Predictive Planning (MPP) and Control (MPC) modules. The two optimization-based solutions share the same problem formulation and model, differing only in horizon length, rate, tuning, and in their open-loop versus closed-loop approach to maximize the effectiveness of their interaction. The work is validated on the fully autonomous open-wheel racecar Superformula EAV-25, with a lap time achieved that is within 2\% of the best human driver reference. The results demonstrate the capability of the solution in driving at the limit of handling, smoothly executing overtaking maneuvers, and quickly reacting to high oversteering conditions to recover the vehicle stability.

\end{abstract}

\section{INTRODUCTION}

Research on motion planning and control of autonomous sportcars and racecars at their physical limit raised an increased interest in the last decade. Beyond works on accurate path tracking, a particular focus has been given on autonomous drifting \cite{peterson,kobayashi,Goh02102024,weber,djeumou}, showcased as a challenging maneuver to be tracked autonomously. Limited works on this topic include transient maneuvers or adapting to unexpected conditions during the drifting \cite{Goh02102024,weber}. Furthermore, they do not focus on demonstrating particular capability of driving-style switching or domain shifting, which could be an important feature for a superhuman autonomous driver on normal roads, taking action in edge scenarios where normal drivers usually fail \cite{kegelman2018learning, heesen}.

In \cite{subosits2024autonomoustestdriverhighperformance}, a Reinforcement Learning  (RL)-based agent effectively drives a racing car with varying vehicle setups that produce understeering, oversteering, and balanced behaviours, while still showing faster laptimes than a human driver. As common for RL, the solution is not validated in a real-world platform despite being tested on a high-fidelity simulator. A more practical work based on Model Predictive Control (MPC), demonstrated the ability to stabilize the vehicle at the limits of the tire forces exploiting steering torque measurement and the vehicle sideslip angle estimation \cite{beal}.
Nevertheless, MPC-based solutions rely heavily on a predefined reference trajectory or setpoint embedded in the stage cost function, which inherently limits their efficacy on different tasks \cite{reiter, song}. Even when augmenting the cost function with additional terms to account for a particular condition \cite{lenssen,bertipaglia2023model}, the paradigm remains anchored to reference tracking, necessitating manual or adaptive tuning of weighting matrices to accommodate new objectives \cite{Puigjaner_2025,zarrouki2021weights}.

\begin{figure}[t]
  \centering
    \includegraphics[width=1.\columnwidth, trim=120 130 180 120, clip]{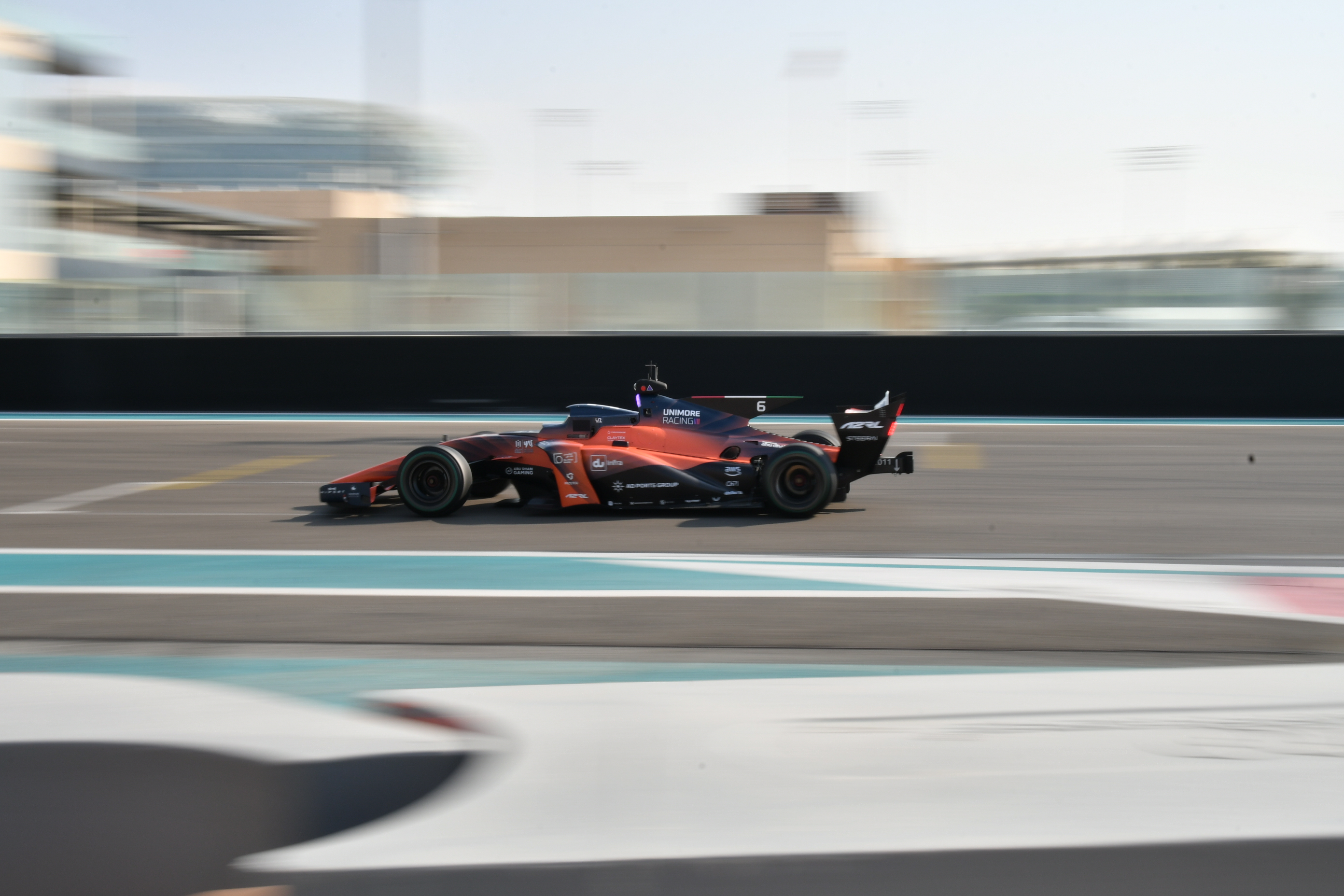}
\caption{Dallara Superformula EAV-25 autonomous racecar, with Unimore Racing livery, at the Yas Marina circuit. \copyright Matteo Villa}
  \label{fig:eav25}
\end{figure}

\begin{figure*}[t!]
\centering
\begin{tikzpicture}[
    node distance=1.5cm and 1.5cm,
    block/.style={rectangle, draw, rounded corners=3pt, fill=blue!5, 
                  minimum width=2.4cm, minimum height=1.1cm, align=center, font=\sffamily\small, line width=0.6pt},
    arrow/.style={-{Stealth[scale=0.9]}, line width=0.7pt, draw=black!80}
]

    
    \node[block, xshift=-3.5cm, fill=blockblue!30] (lp) {Longitudinal\\Planner};
    \node[block, right=of lp, fill=blockblue!30] (mpp) {Model Predictive\\Planning (MPP)};
    \node[block, above=1.8cm of mpp, fill=blockgray!30] (fsm) {Cost Blending\\State Machine};
    \node[block, right=of mpp, fill=blockorange!30] (mpc) {Model Predictive\\Control (MPC)};
    
    \node[block, right=1.5cm of mpc, fill=blockorange!30] (low) {Low Level\\Control};
    \node[above=1.5cm of low] (car) {
        \includegraphics[trim={0 150 100 100}, clip, width=5cm]{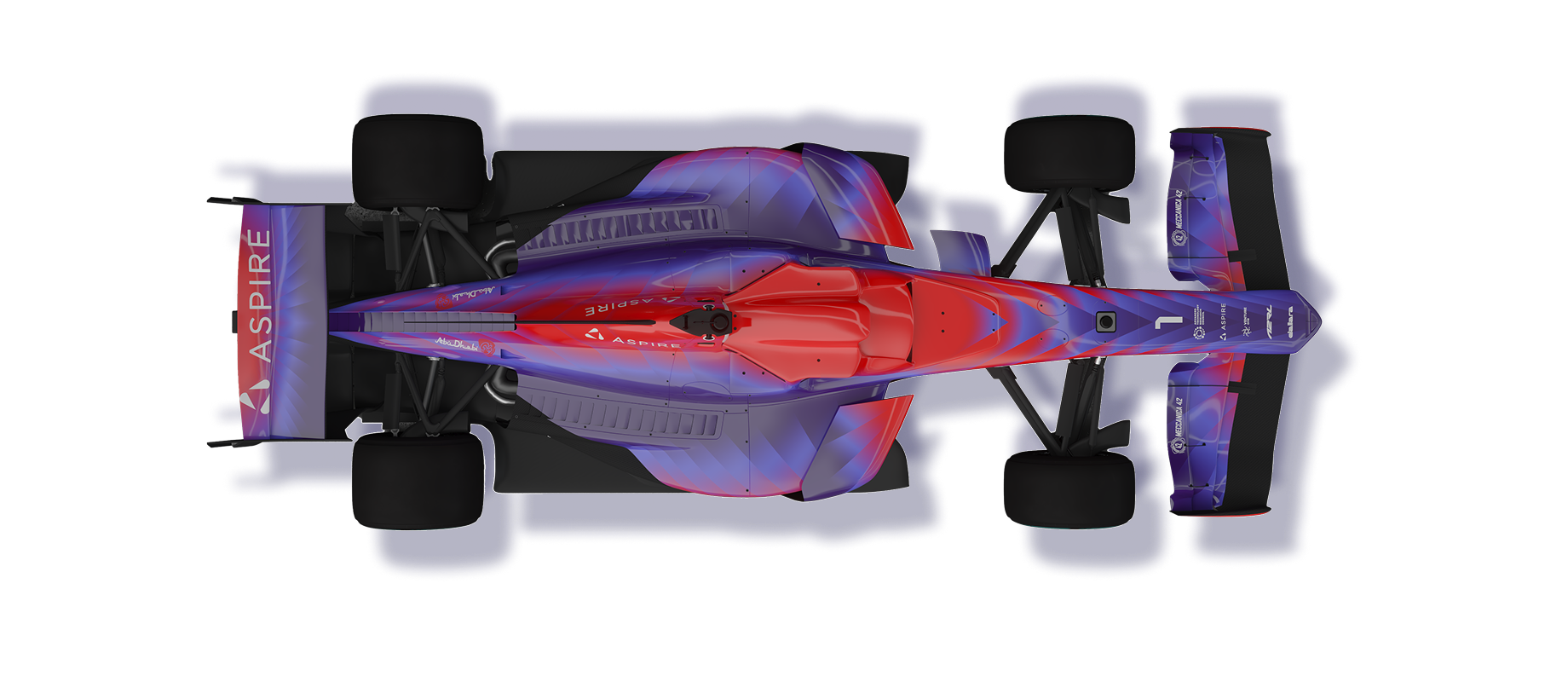}
    };

    
    \draw[arrow] (fsm.south) -- ++(0,-0.7) node[midway, right, font=\small, xshift=2pt] {$weights$} -| (mpp.north);
    \draw[arrow] (fsm.south) -- ++(0,-0.7) -| (mpc.north);
    
    \draw[arrow] ([yshift=0.15cm]lp.east) -- node[above, font=\small] {$v_x$} ([yshift=0.15cm]mpp.west);
    \draw[arrow] ([yshift=-0.15cm]mpp.west) -- node[below, font=\small] {$path$} ([yshift=-0.15cm]lp.east);
    \draw[arrow] (mpp.east) -- node[above, font=\small] {$path$} node[below, font=\small] {$v_x profile$} (mpc.west);
    
    \draw[arrow] (mpc.east) -- ++(0.5,0) |- node[pos=0.75, above, font=\small] {$\delta$} ([xshift=+0.8cm]car.west);
    
    \draw[arrow] (mpc.east) -- ++(0.5,0) |- node[pos=0.75, above, font=\small] {$a_x$} (low.west);
    
    \draw[arrow] (low.north) -- node[right, font=\small, align=center] {$throttle$\\[0.5ex]$brakes$\\[0.5ex]$gear$} (car.south);

\end{tikzpicture}
\caption{System Architecture Overview. In blue the nodes related to the Planning module. In orange, the Control-related modules. In the top, the Cost Blending State Machine that selects the weights for the MPP and MPC cost functions. $\delta$ represents the steering angle. $a_x$ the longitudinal acceleration to be tracked by the low level control.}
\label{fig:system_overview}
\end{figure*}
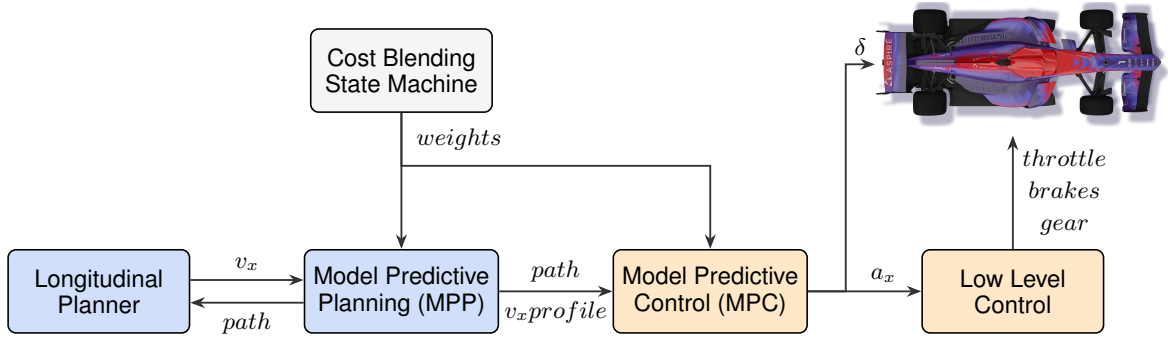

MPC capable of balancing path tracking, stability, and obstacle avoidance is demonstrated in \cite{lenssen,bertipaglia2023model}. In particular, the authors include additional terms to the function cost, related to the difference between a planned yaw rate and the actual one, and a user-defined safety distance between the ego vehicle and the obstacle. Promising results are demonstrated also with varying friction and model mismatch, although they are limited to simulation on double-lane change maneuvers. Furthermore, the approach does not scale efficiently to additional driving modes, such as a comfort mode that maintains accurate path tracking while prioritizing smoothness through jerk minimization, as each new mode requires distinct cost terms, increasing the potential for conflicts and undesirable interactions among cost components.
In \cite{zarrouki2021weights}, the weights used by an MPC are updated online by an RL agent with training policies focused on a certain driving mode between path tracking and comfort, making it a more extensible solution for new driving modes. 

In contrast to prior state-of-the-art methods, this paper presents a more hierarchical, scalable, and modular approach demonstrating effectiveness in switching between driving modes while dynamically managing inter-mode priorities. A state machine detects the driving context and chooses the cost function weights to be used by the planning (MPP) and the control (MPC) modules. Additional contributions are summarized as follows:

\begin{itemize}
    \item Demonstration on a fully autonomous open-wheel racecar (Dallara Superformula EAV-25, shown in Figure \ref{fig:eav25}) achieving professional-level performance, with laptimes within 2\% of the best human reference.
    \item This paper addresses vehicle control beyond handling limits under extreme racing conditions. While previous controllers handle countersteering \cite{beal}, they lack the focus on critical dynamics—such as high sideslip and model mismatch.
    \item A dual-MPC architecture (MPPC), comprising a planning MPC, for the remainder of the paper it will be named MPP,  and a control MPC formulated with the same vehicle model but different tuning and an open-loop versus closed-loop approach, ensures coherent interaction between modules; this contrasts with literature solutions that typically employ simpler or less accurate models in planning relative to control \cite{Stahl_2019,ogretmen2022,ogretmen2024sampling}.
\end{itemize}
In Section \ref{sec:sys_overview}, an overview of the complete system is given, motivating the design choices. The MPPC formulation is presented in Section \ref{sec:mppc}, as well as information about the vehicle model used. A particular explanation of the driving context-based Cost Blending State Machine is given in Section \ref{subsec:cost_weighting}. In Section \ref{sec:results}, the validation setup used and the experimental results on the three main driving conditions (nominal driving, overtaking and oversteering) are described. Conclusion and future work are presented in Section \ref{sec:conclusions}.

\section{SYSTEM OVERVIEW}
\label{sec:sys_overview}

Several planning-control architectures are proposed in the literature: one approach simplifies the planning layer, relying on a robust controller to handle disturbances; another simplifies the control layer, trusting a complex planner to generate physically feasible references \cite{Betz_2022}.

In our framework, we adopt a hybrid solution: the same vehicle model complexity is used for both MPP (trajectory generation) and MPC (trajectory tracking) by sharing the same model and parameters. This ensures that:
\begin{itemize}
\item The MPP generates physically accurate and feasible references for 
all maneuvers, fully exploiting the model's capabilities, thus filtering
unfeasible reference from oversimplified higher-level layer.
\item The MPC can focus on aggressive tracking with reduced risk of 
infeasible commands, since the reference it receives is designed to be 
executable, still being able to manage high-dynamic scenarios.
\end{itemize}
This separation allows the planner to be tuned for smoothness and 
stability, producing high-quality references, while the controller can be 
tuned for maximal tracking performance, knowing that the input trajectory 
is already physically consistent.

Despite not being the focus of the paper, the overall planning and control scheme, depicted in Figure \ref{fig:system_overview}, in which the main contributions of the paper are included, is briefly described in this section. The core modules of the work are presented more in detail in Section \ref{sec:mppc} and Section \ref{subsec:cost_weighting}.

\subsection{Longitudinal Planner}
\label{subsec:long_planner}
Producing different speed profiles along the optimal path directly within the MPP is neither straightforward nor flexible. In addition, a good initial guess is required to ensure a safe and efficient optimization within the MPP.

To address this, a dedicated longitudinal planner computes the velocity profile along the path through a forward–backward integration scheme, enforcing lateral and longitudinal acceleration constraints \cite{ur11}. Its role is to translate normalized friction-ellipse (dynamic GG diagram) performance requests into velocity targets for the MPP. The GG bounds, i.e., the maximum achievable accelerations, are derived from the maximum axle forces obtained by the friction law, with the tangential force peak of each axle calculated from the Pacejka peak coefficients\cite{pacejka1992magic}.

\subsection{Model Predictive Planner (MPP)}
\label{subsec:mpp}

\begin{table}[t]
\centering
\caption{Main objectives of the MPP.}
\label{tab:mpp_objectives}
\begin{tabularx}{\columnwidth}{l>{\small\arraybackslash}X}
\toprule
\textbf{Objective} & \textbf{Description} \\
\midrule

Optimal trajectory adjustment & Optimize the offline computed global path 
and longitudinal planner speed profile, based on the current vehicle model 
or newly identified dynamics. \\

\hline

Optimal trajectory deviation & Compute a new physics-based trajectory for 
specific maneuvers, such as overtaking, line switch and obstacle avoidance, 
respecting the collision-free navigable area. \\

\bottomrule
\end{tabularx}
\end{table}

The Model Predictive Planner (MPP) operates in open-loop at $20\,\mathrm{Hz}$ with a sampling time $\Delta t = 0.04\,\mathrm{s}$, over an horizon of $N=150$ steps ($T=6\,\mathrm{s}$).
In an open-loop operation, the planner computes the path assuming that the previous iteration was executed perfectly, without using vehicle feedback to update its initial state $\hat{\text{x}}$. 
Since the actual longitudinal position $s$ of the vehicle may differ from the planned one, the planner reprojects the previous solution onto the current longitudinal position and uses it as the initial condition for the new MPP execution.

The planner generates smooth and feasible paths within the track margins, 
which coincide with the physical boundaries under nominal conditions.
In multi-agent scenarios, a specific module updates 
the track margins to account for opponent vehicles, enforcing predefined 
lateral and longitudinal safety distances \cite{toschi}.
As a result, the planner is required to generate trajectories that 
lie within the resulting obstacle-free navigable region, while simultaneously 
satisfying the dynamic constraints imposed by the optimization problem.
The main objectives of the MPP are detailed in Tab.\ref{tab:mpp_objectives}.

The open-loop design was adopted for the following reasons:
\begin{itemize}
    \item In a closed-loop formulation, the planner recomputes the trajectory
    starting from the full ego vehicle states and position at each iteration, artificially
    resetting the controller tracking error and reducing the effectiveness
    of feedback.
    In contrast, the open-loop approach allows the controller to naturally
    handle tracking errors.
    \item It provides a consistent and intuitive definition of tracking errors
    in the vehicle reference frame, such as the lateral deviation from the
    reference path (in this case, generated by the MPP, being the planner module), for evaluation of controller performance.
    \item It guarantees a stable and continuous reference trajectory for the
    controller, simplifying both the planning and control optimization problems,
    and allowing the controller (operating on a shorter horizon but at a higher
    frequency) to handle disturbances effectively.
    \item 
    The open-loop MPP significantly simplifies tuning and validation. Since 
    its behavior is independent of vehicle feedback, it can be evaluated in 
    isolation under controlled simulation conditions, enabling reproducible 
    tests and more systematic calibration using clearly defined scenarios.
\end{itemize}
The loop is closed for an iteration when specific conditions occur, for example:
\begin{itemize}
    \item Lateral, heading, or longitudinal errors exceeding a predefined 
    threshold.
    \item A change in objective, e.g., line switching.
\end{itemize}

\begin{table}[t]
\centering
\caption{Main objectives of the MPC.}
\label{tab:mpc_objectives}
\begin{tabularx}{\columnwidth}{l>{\small\arraybackslash}X}
\toprule
\textbf{Objective} & \textbf{Description} \\
\midrule
Path-following & Computes the optimal control action to follow the planned 
path and speed.\\
\hline

Vehicle dynamics stability & Compensate for feedback deviations in driving at the limit and potentially critical scenarios, such as oversteer. \\
\bottomrule
\end{tabularx}
\end{table}

\subsection{Model Predictive Controller (MPC)}
\label{subsec:mpc}
The MPC operates in closed-loop at $100\,\mathrm{Hz}$ with 
the same sampling time used for the MPP ($\Delta t = 0.04\,\mathrm{s}$), over $N=64$ 
steps (time horizon $T=2.6\,\mathrm{s}$). 
The initial state is updated using vehicle feedback, filtered through 
localization and state estimation modules.
The main objectives of the MPC are summarized in Tab.~\ref{tab:mpc_objectives}.

\noindent
The closed-loop design follows this rationale:
\begin{itemize}
  \item Controller weights are primarily focused on tracking and stability, 
  assuming the planner already provides a physically feasible, optimal 
  trajectory for the current vehicle model and track conditions.
  \item Handling model mismatches, compensating errors, and ensuring path-following, are entirely the controller's responsibility, rather 
  than the planner's. Non-nominal scenarios (e.g., overtaking or off-track driving) should be handled by the planner, producing an adequate new trajectory as a reference for the MPC.
  \item Track constraints are softly enforced, allowing the controller to 
  prioritize stability during critical maneuvers.
\end{itemize}
The longitudinal acceleration $a_x$ from the MPC's output, is tracked by a low-level longitudinal control module formed by a feedforward component and a PID for feedback compensation, producing final throttle and brake actuations to the vehicle \cite{ur11}. The gear selection is produced by a separate component of the low-level control matching speed, engine rpm, and yaw rate \cite{ur11}.
\subsection{Cost Blending}
\label{subsec:cost_blending}
A single cost function may fail to capture the variety of scenarios 
encountered in autonomous racing, as different maneuvers can have distinct 
objectives. Using a single cost set can either provide a compromise across all scenarios or be overfit to a specific one. To address this limitation, we define multiple cost sets targeting specific planning and control goals. 
A cost-blending state machine dynamically selects the appropriate MPPC cost 
set based on the current driving scenario, as summarized in 
Tab.~\ref{tab:cost-blend}.
\begin{table}[h!]
\centering
\caption{MPPC Cost-Blending States and Activation Criteria. $d_\text{opponent} < d_\text{thresh}$ represents the distance check from the opponent. $\alpha_f$ is the front slip angle, $\alpha_r$ is the rear slip angle, and $U_\text{th}$ is a (positive) threshold for the understeer angle value representing a stability metric of the vehicle \cite{guiggiani2014science}.}\label{tab:cost-blend}
\resizebox{\columnwidth}{!}{%
\begin{tabular}{ccc} 
\toprule
\textbf{State} & \textbf{Description} & \textbf{Activation Condition} \\
\midrule
Nominal & Baseline behaviour. & Default state \\
Line switch & Transition between reference lines. & $\text{High-Level Request}$ \\
Overtaking & Attack/defence logic triggered by opponent's interaction. & $d_\text{opponent} < d_\text{thresh}$ \\
Oversteer & Stability-recovery mode. & $|\alpha_f|-|\alpha_r| < -U_\text{th}$ \\
Out of track & Vehicle leaves track limits. & $|n| > n_\text{track}$ \\
\bottomrule
\end{tabular}%
}
\end{table}
The state machine provides smooth, context-aware updates of the MPPC cost.
Cost blending is performed using sigmoid-based interpolation, ensuring 
gradual transitions between different cost sets without discontinuities.

\begin{figure*}[t!]
\centering
\begin{tikzpicture}[
    node distance=1.3cm, 
    every node/.style={
        rectangle, 
        draw, 
        rounded corners=3pt, 
        minimum width=2.2cm, 
        minimum height=1.1cm, 
        align=center,
        line width=0.8pt,
        font=\small\sffamily
    },
    arrow/.style={
        -{Stealth[scale=1.0]}, 
        line width=0.7pt, 
        draw=black!80
    }
]

    \node (nominal) [fill=blockgray!30]{Nominal};
    \node (line) [right=of nominal, fill=blockblue!30] {Line Switch};
    \node (overtaking) [right=of line, fill=blockblue!30] {Overtaking};
    \node (oob) [right=of overtaking, fill=blockgray!30] {Out of Bounds};
    \node (oversteer) [right=of oob, fill=blockorange!30]{Oversteer};

    \draw [arrow] ([yshift=2mm]nominal.east) -- ([yshift=2mm]line.west);
    \draw [arrow] ([yshift=-2mm]line.west) -- ([yshift=-2mm]nominal.east);
    \draw [arrow] (line.east) -- (overtaking.west);
    \draw [arrow] (overtaking.east) -- (oob.west);
    \draw [arrow] (oob.east) -- (oversteer.west);

\draw [{Stealth[scale=1.0]}-{Stealth[scale=1.0]}, line width=0.7pt, draw=black!80] (overtaking.north) -- ++(0, 1.0) -| ([xshift=4mm]nominal.north);
    \draw [{Stealth[scale=1.0]}-{Stealth[scale=1.0]}, line width=0.7pt, draw=black!80] (oob.north) -- ++(0, 1.4) -| ([xshift=1mm]nominal.north);

    \draw [{Stealth[scale=1.0]}-{Stealth[scale=1.0]}, line width=0.7pt, draw=black!80] 
        (nominal.south) -- ++(0, -1.6) -| ([xshift=6mm]oversteer.south);
    
    \draw [arrow] (line.south) -- ++(0, -1.1) -| ([xshift=2mm]oversteer.south);
    \draw [arrow] (overtaking.south) -- ++(0, -0.6) -| ([xshift=-2mm]oversteer.south);
    \draw [arrow] (oob.south) -- ++(0, -0.3) -| ([xshift=-6mm]oversteer.south);

\end{tikzpicture}
\caption{Cost Blending State Machine scheme. In blue the states related to driving modes for MPP; in orange the ones related to MPC and in grey the ones used by both the modules.}
\label{fig:fsm_bidir}
\end{figure*}
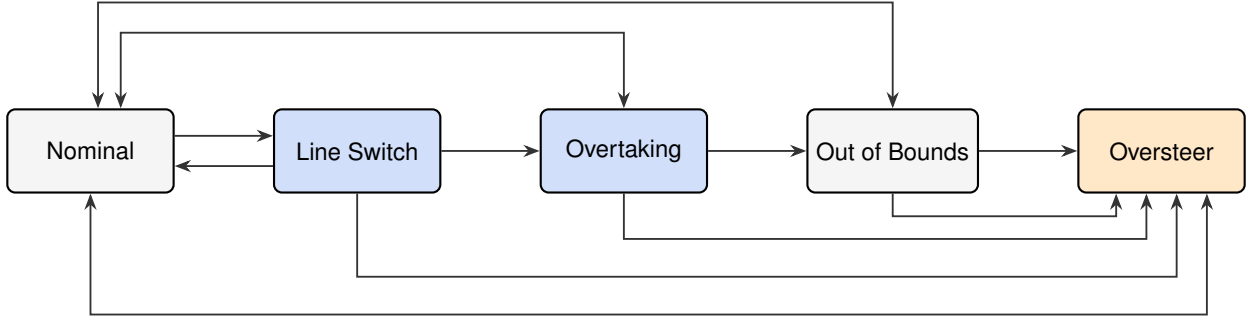

\section{MPPC FORMULATION}
\label{sec:mppc}
The optimization problem for MPP and MPC is solved using a custom Sequential Quadratic Programming (SQP) framework that takes advantage of the HPIPM library \cite{hpipm}, a high-performance quadratic programming solver specifically designed for MPC applications, together with CppADCodeGen for automatic differentiation \cite{ur11}. These tools are essential to achieve real-time computational performance.

Although the following subsections give an overview of the main parts of the formulation, further details are explained in \cite{Musiu}.
\subsection{VEHICLE MODEL}
\label{subsec:model}
As a model for MPP and MPC, an enhanced nonlinear single-track model incorporating the Pacejka Magic Formula tire model is used \cite{Musiu}. To minimize the computational burden while maintaining accuracy and to make the model suitable for the SQP, it is linearized over the previous horizon prediction.
The state is $\text{x} = [\, s;\, n;\, \mu;\, v_x;\, v_y;\, r;\, \delta;\, a_x\, ]$, where $s$ denotes the progress along the path, $n$ the lateral deviation, and $\mu$ the local heading angle with respect to the global path.
The input vector is given by $\text{u} = [\, \dot{\delta};\, {J}_x\,]$, where $\dot{\delta}$ denotes the steering rate and ${J}_x$ is the longitudinal jerk ($\frac{\delta a_x}{\delta t}$).
The full motion equations are written following the formulation in \cite{Musiu}. 

The model $f_t^d(\text{x}_k, \text{u}_k)$ is discretized using a 4th-order Runge-Kutta method and formulated in curvilinear coordinates, where $\hat{\text{x}}$ is the current curvilinear state.
At low speed ($v_x \leq 8 \frac{m}{s}$) the dynamic model is replaced with a kinematic formulation while retaining the same state-space structure. This approach avoids numerical instabilities while providing a reliable approximation of vehicle behavior. In particular, as the speed increases, an exponential blending function smoothly transitions the model from kinematic to dynamic, ensuring that all relevant vehicle dynamics are accurately captured.

\subsection{Problem Formulation}
\label{subsec:mpc_problem}
The optimization problem is formulated as:
  \begin{equation}\label{eq:mpc_opt}
      \begin{aligned}
        \min_{\text{X}, \text{U}} &\quad \sum_{k=0}^{T} J_\text{MPC}(\text{x}_k, \text{u}_k) \\
        \text{s.t.} &\quad \text{x}_0 = \hat{\text{x}}, \\
        &\quad \text{x}_{k+1} = f_t^d(\text{x}_k, \text{u}_k), \\
        &\quad \text{x}_k \in \text{X}_{\text{track}}, \quad 
               \text{x}_k \in \text{X}_{\text{ellipse}}, \\
        &\quad \text{a}_k \in \boldsymbol{\text{A}}, \quad 
               \text{u}_k \in \boldsymbol{\text{U}}, \quad 
               k = 0, \dots, T, \\
      \end{aligned}
  \end{equation}
\noindent
where $\text{X}=[\text{x}_0, \ldots, \text{x}_N]$ and $\text{U}=[\text{u}_0, \ldots, \text{u}_N]$ represent the set of state and input vectors over the entire horizon. $X_{ellipse}$ represents the tire friction ellipse constraints, and $X_{track}$ represents a track constraint on the lateral deviation $n$ ensuring that the trajectory stays on the drivable area generated by \cite{toschi}. The inputs $a = [\delta; J_x]$ and their rate of change $u$ are constrained using box constraints $\bm{A}$ and $\bm{U}$.

The objective function is formulated as:
\begin{equation}\label{eq:cost_f}
  J_{\textit{MPC}}(\text{x}_k, \text{u}_k) = 
  \text{x}_k^{T} Q \text{x}_k + 
  \text{u}_k^{T} R \text{u}_k + 
  B(\text{x}_k)
\end{equation}
\noindent
where $Q$ and $R$ denote the state and input weighting matrices, respectively.
The state weighting matrix includes path-following terms $q_n$ and $q_\mu$,
a velocity tracking weight $q_{vx}$, and a yaw rate weight $q_r$ to enhance
stability and dampen oscillations. The regularization term 
$\text{u}^T R \text{u}$ penalizes input variations and promotes smoother 
control actions. Finally, the term $B(\text{x}_k) = q_{\beta} \beta^2$ 
penalizes the sideslip angle ($\beta$), further contributing to vehicle stability when operating near the limits of tire friction.

\section{DRIVING CONTEXT-BASED COST WEIGHTING}
\label{subsec:cost_weighting}

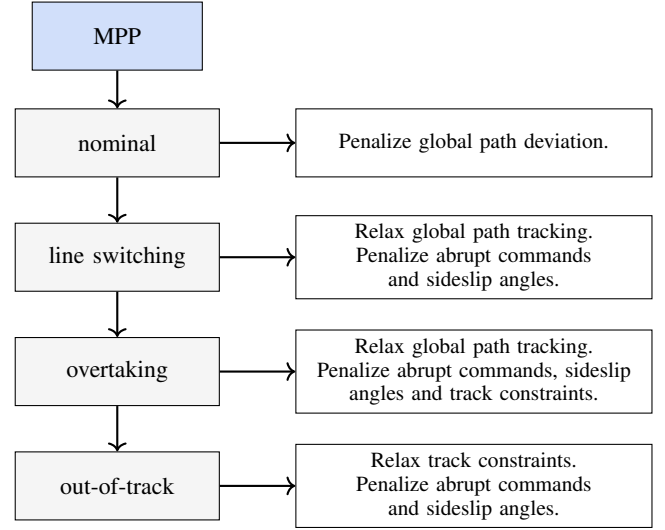
\begin{figure}[]
\centering
\begin{tikzpicture}[node distance=1cm, auto, scale=0.9, every node/.style={scale=0.9}]

\node[draw, fill=blockblue!30, minimum width=2.5cm, minimum height=1cm] (mpp) {MPP};

\node[draw, fill=blockgray!30, minimum width=3cm, minimum height=1cm, below=0.5cm of mpp] (sw1) {nominal};
\node[draw, fill=blockgray!30, minimum width=3cm, minimum height=1cm, below=0.6cm of sw1] (sw2) {line switching};
\node[draw, fill=blockgray!30, minimum width=3cm, minimum height=1cm, below=0.6cm of sw2] (sw3) {overtaking};
\node[draw, fill=blockgray!30, minimum width=3cm, minimum height=1cm, below=0.6cm of sw3] (sw4) {out-of-track};

\node[draw, minimum width=5cm, minimum height=1cm, right=1cm of sw1, text width=5cm, align=center, font=\small] (exp1) {Penalize global path deviation.};
\node[draw, minimum width=5cm, minimum height=1cm, right=1cm of sw2, text width=5cm, align=center, font=\small] (exp2) {Relax global path tracking. \\ Penalize abrupt commands and sideslip angles.};
\node[draw, minimum width=5cm, minimum height=1cm, right=1cm of sw3, text width=5cm, align=center, font=\small] (exp3) {Relax global path tracking. \\ Penalize abrupt commands, sideslip angles and track constraints.};
\node[draw, minimum width=5cm, minimum height=1cm, right=1cm of sw4, text width=5cm, align=center, font=\small] (exp4) {Relax track constraints. \\ Penalize abrupt commands and sideslip angles.};

\draw[->, thick] (mpp.south) -- (sw1.north);

\draw[->, thick] (sw1.south) -- (sw2.north);
\draw[->, thick] (sw2.south) -- (sw3.north);
\draw[->, thick] (sw3.south) -- (sw4.north);

\draw[->, thick] (sw1.east) -- (exp1.west);
\draw[->, thick] (sw2.east) -- (exp2.west);
\draw[->, thick] (sw3.east) -- (exp3.west);
\draw[->, thick] (sw4.east) -- (exp4.west);

\end{tikzpicture}
\caption{Planner process: vertical switch stages with corresponding cost explanations.}
\label{fig:planner_switch_vertical}
\end{figure}

In Figure \ref{fig:fsm_bidir}, a scheme of the Cost Blending State Machine (CB-SM) shows the different contexts (Nominal Driving, Line Switch, Overtaking, Out-of-Track, and Oversteer) and the allowed transitions. MPP and MPC have separate instances of the CB-SM. In particular, the planner implements the cost blending among four main objectives: 
Nominal, Line Switching, Overtaking, and Out-of-Track scenarios, as detailed in Fig.\ref{fig:planner_switch_vertical}.
These scenarios are closely related to trajectory generation, hence deemed planner tasks, while tracking-error compensation (both in nominal and high-dynamics conditions) is handled by the controller.
In the MPC, cost-blending is implemented between two main objectives: Nominal and 
Oversteer, as detailed in Fig.\ref{fig:controller_switch_vertical}.
These scenarios are tightly linked to trajectory tracking and are therefore 
considered controller-level tasks.

As illustrated in Figure \ref{fig:fsm_bidir}, the allowed transitions define a priority hierarchy among driving contexts. Given that the Oversteer condition poses the most significant threat to vehicle stability, the architecture permits direct transitions from any state to the Oversteer state. Conversely, for all other transitions, the system must first satisfy the exit conditions specified in Table \ref{tab:cost-blend} to revert to the Nominal state before switching to a new operating mode.

\subsection{Cost Blending}
\label{subsec:cost_blending2}
Depending on the current state of the CB-SM, the output cost might depend on the sigmoid function described as 
\[
\text{sigmoid}(x) = m + \Bigl( (M - m) \cdot \left( 1 - \frac{1}{1 + e^{\tfrac{c - x}{\tau}}} \right) \Bigr)
\]
where:
\begin{itemize}
    \item $m$: minimum value 
    \item $M$: maximum value 
    \item $c$: center value
    \item $\tau$: slope 
\end{itemize}

Then, the actual multiplier is computed by:
\[
\text{mult}(x) = 1 - min(\text{sigmoid}(x), \text{negativeSigmoid}(x)) 
\]
where $\text{negativeSigmoid}(x)$ is a sigmoid, which has the maximum and minimum values swapped.\\
Finally, the new cost value is calculated by imposing an interval $[a,b]$ in which there should exist.
\[
\text{cost}(x) = min(a,b) + \text{mult}(x)|a-b|  
\]

The magnitude of the line switch cost, depends on the current vehicle's lateral displacement with respect to the selected racing line (i.e., the further the vehicle is from the racing line, the more the cost will tend to the line switch cost). The same process also goes for the oversteering cost set on the understeer degree threshold $U_{th}$. Instead, regarding the overtaking and out of bounds costs, the switch is not regulated (and smoothed) by the sigmoid function, but it is discrete (i.e., an on/off mechanism).
Each switching is triggered (and regulated, if it is the case) by monitoring a corresponding set of states of the vehicle (e.g., for line switching the Frenet coordinates).

\subsection{Tuning}
\label{subsec:tuning}
\begin{figure}[]
\centering
\begin{tikzpicture}[node distance=1cm, auto, scale=0.9, every node/.style={scale=0.9}]

\node[draw, fill=blockorange!30, minimum width=2.5cm, minimum height=1cm] (mpc) {MPC};

\node[draw, fill=blockgray!30, minimum width=3cm, minimum height=1cm, below=0.5cm of mpc] (sw1) {nominal};
\node[draw, fill=blockgray!30, minimum width=3cm, minimum height=1cm, below=0.6cm of sw1] (sw2) {oversteer};

\node[draw, minimum width=5cm, minimum height=1cm, right=1cm of sw1, text width=5cm, align=center, font=\small] (exp1) {Penalize local path deviation, sideslip angles, and abrupt commands.};
\node[draw, minimum width=5cm, minimum height=1cm, right=1cm of sw2, text width=5cm, align=center, font=\small] (exp2) {Relax local path tracking and steering derivative. \\ Penalize sideslip angle \\and heading error.};

\draw[->, thick] (mpc.south) -- (sw1.north);

\draw[->, thick] (sw1.south) -- (sw2.north);

\draw[->, thick] (sw1.east) -- (exp1.west);
\draw[->, thick] (sw2.east) -- (exp2.west);

\end{tikzpicture}
\caption{Controller process: vertical switch stages with corresponding cost explanations.}
\label{fig:controller_switch_vertical}
\end{figure}
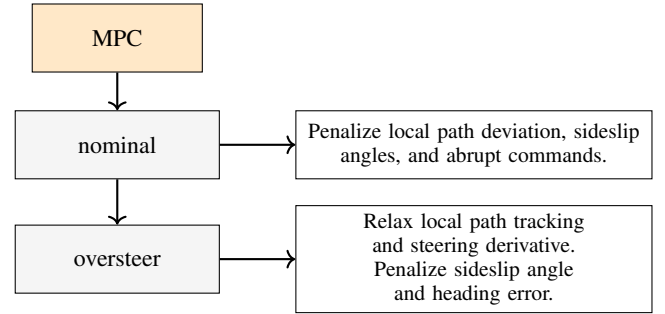

Following the weighting logic for each different context, as depicted in Figure \ref{fig:planner_switch_vertical} and Figure \ref{fig:controller_switch_vertical}, a dedicated tuning has been found. In particular, the tuning for the Overtaking condition has been considered acceptable also for the Line Switching, therefore in Table \ref{tab:costs} the latter is not depicted. It can be noticed that the main differences are on the path tracking $q_n$ and side slip angle $q_{\beta}$ terms for all the conditions. The Out of Track condition highlights the need for relaxation for the track soft constraint. In this situation, the optimization process may otherwise generate abrupt control actions aimed at returning the vehicle to the track, potentially causing instability. This behavior is particularly critical, as being out of track already indicates a substantial performance mismatch or a hazardous operating condition.

While tuning the MPP-related costs is relatively straightforward, focusing primarily on generating smoother control actions, the MPC configuration for the Oversteer condition is more nuanced.
Beyond reducing the lateral error weight ($q_n$) and increasing the sideslip penalty ($q_{\beta}$), the heading error weight ($q_{\mu}$) is also increased. This adjustment is strategic, as prioritizing path alignment during oversteer is beneficial to regain vehicle stability. Furthermore, we increase the soft constraints on track limits as a safety measure since the oversteer could occur also during overtaking maneuvers or when the vehicle is near the physical track boundaries.

\begin{table}[!h]
\centering
\caption{
High-level weights tuning of each Driving Condition}
\label{tab:costs}
\renewcommand{\arraystretch}{.8} 

\begin{subtable}{\columnwidth}
\centering
\caption{Planner weights: OT represents Overtaking; OoT stands for Out of Track}
\begin{tabularx}{\columnwidth}{*{10}{>{\scriptsize\centering\arraybackslash}X}}
\toprule
 & $q_n$ & $q_{\mu}$ & $q_{v_x}$ & $q_r$ & $q_{\beta}$ & $r_{d\delta}$ & $r_{J_x}$ & $s_{track}$ & $s_{tires}$ \\
\midrule
Nominal & ++ & ++ & ++ & + & +++ & +++ & ++ & +++ & +++ \\
OT & + & +++ & ++ & ++ & ++++ & ++++ & +++ & ++++ & +++ \\
OoT & + & +++ & ++ & ++ & ++++ & ++++ & +++ & + & +++ \\
\bottomrule
\end{tabularx}
\end{subtable}

\vspace{0.2cm} 

\begin{subtable}{\columnwidth}
\centering
\caption{Controller weights}
\begin{tabularx}{\columnwidth}{*{10}{>{\scriptsize\centering\arraybackslash}X}}
\toprule
 & $q_n$ & $q_{\mu}$ & $q_{v_x}$ & $q_r$ & $q_{\beta}$ & $r_{d\delta}$ & $r_{J_x}$ & $s_{track}$ & $s_{tires}$ \\
\midrule
Nominal & ++++ & +++ & +++ & + & +++ & ++ & + & + & + \\
Oversteer & + & +++++ & + & ++ & +++++ & + & ++ & ++ & + \\
\bottomrule
\end{tabularx}
\end{subtable}

\vspace{0.5cm} 

\footnotesize
\begin{tabularx}{\columnwidth}{lX}
\textbf{Symbol} & \textbf{Meaning} \\
\hline
++++   & Strong prioritization \\
+++    & Medium \\
++     & Low \\
+   & Weak prioritization \\
\end{tabularx}

\end{table}

\section{RESULTS}
\label{sec:results}

\subsection{Validation Setup}
\label{subsec:validation_setup}
The proposed solution is tested on a full-scale open-wheel racecar, the Dallara Superformula EAV-25, equipped with 3 LiDARs, 4 RADARs, 6 cameras, a GNSS module, 2 IMUs, an optical sensor for sideslip angle measurement, and other classical sensors. The states necessary for the MPP and MPC are produced by dedicated state estimation and localization modules based on a set of Extended Kalman Filters (EKF) exploiting estimated odometries produced using the sensor data.
A Sensor Fusion-based approach is implemented for the Detection and Tracking of other vehicles, where their future movements are predicted through a dedicated Motion Forecasting module.

The algorithms are executed onboard on a rugged platform equipped with an AMD EPYC 7003 CPU (up to 3.67GHz) and 64GB RAM. The MPP and MPC run within the chosen frequencies (20Hz and 100Hz) with an average execution time of 15ms and 5ms, respectively.

A high-fidelity simulation pipeline is also used to better demonstrate the upside of the dedicated Overtaking costs \cite{lambertini}.
In the following, the effectiveness in the three main driving scenarios of the CB-SM is examined.
\subsection{Nominal Driving Scenario}
\label{subsec:nominal}

The autonomous agent achieved a lap time of 58.76 s, within 2\% of the 57.57 s recorded by a former Formula 1 driver. The comparison was performed during the Abu Dhabi Autonomous Racing League (A2RL) Season 2, using the same vehicle, identical vehicle setup, and under the same track conditions (Yas Marina Circuit, North Configuration).

\begin{figure}[]
  \centering
    \includegraphics[width=1.\linewidth, trim=0 0 0 0, clip]{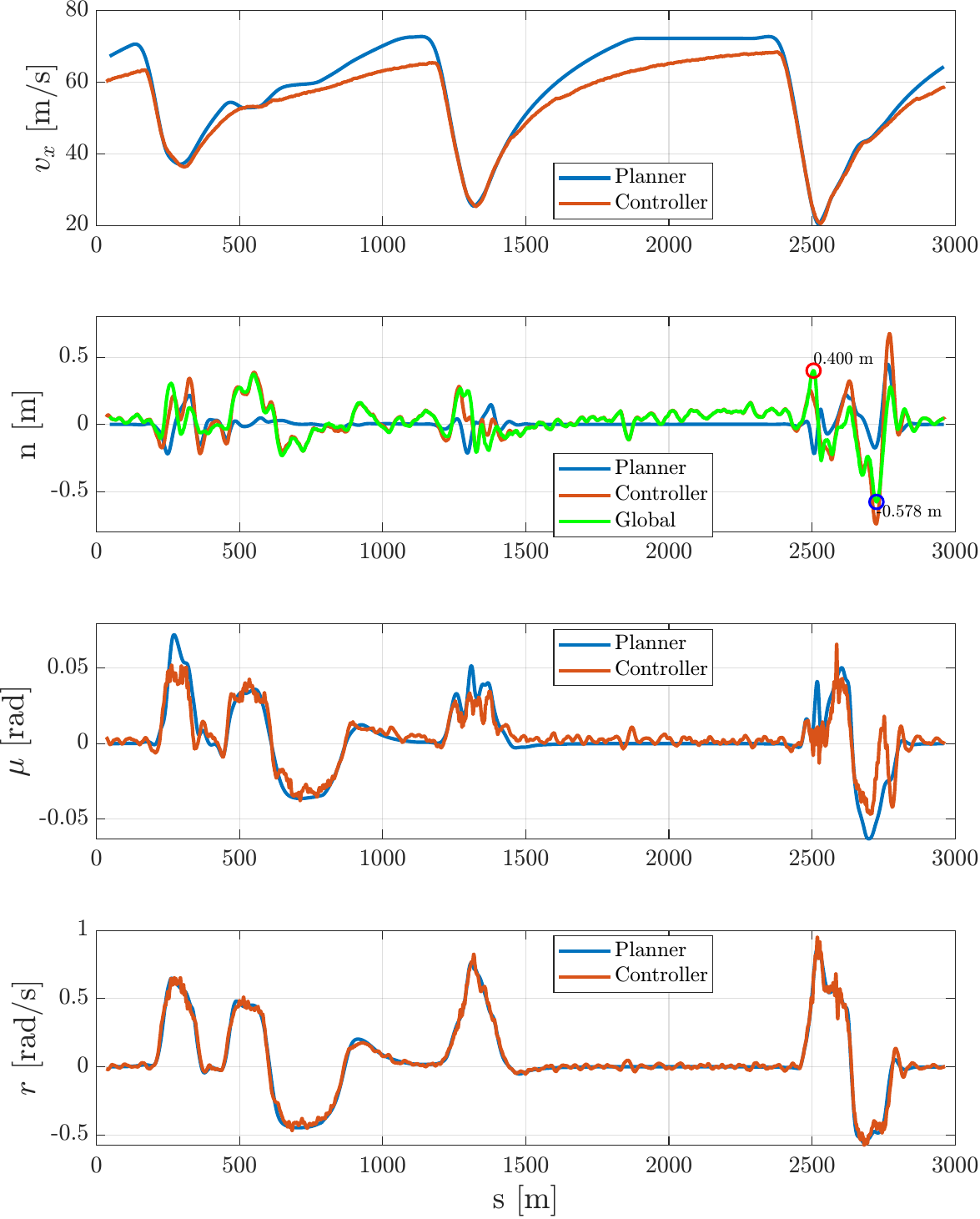}
\caption{Fastest lap overview. From top to bottom: longitudinal velocity ($v_x$), lateral deviation ($n$), heading error ($\mu$), and yaw rate ($r$) along the lap progression ($s$). The green line highlights the differences between planner and controller, representing the effective tracking error.
}
  \label{fig:fast_lap}
\end{figure}

In Figure~\ref{fig:fast_lap}, the path-tracking performance of the MPC is compared with the reference signals planned by the MPP.

Good agreement is observed in the longitudinal velocity ($v_x$), with cornering speeds accurately tracked. Minor discrepancies appear in Turn 1 (s $\approx 300$ m), where the turbo-lag effect, not modeled in the controller layer, creates a mismatch during the corner-exit phase. Larger mismatches are visible in the high-speed sections (e.g., s $\approx 1000$ m and s $\approx 2000$ m), mainly due to a deliberate slight overestimation of engine performance in the longitudinal model. This choice ensures (i) jerk-free reference for the controller during braking and (ii) sustained full-throttle conditions along straight sections thanks to the velocity gap created between planner and controller.
The yaw rate ($r$) is accurately tracked as well.

The lateral deviation ($n$) remains consistently within 40 cm. However, in Turn 8 ($s \approx 2700$ m), an outlier of $n \approx 0.6$ m is observed. This tracking error is caused by the unmodeled three-dimensional road profile in the controller: at the turn apex, a bump induces an additional vehicle displacement due to the increased vertical load on the axles, which in turn affects the lateral forces.

\subsection{Overtaking Scenario}
\label{subsec:overtaking}

\begin{figure}[!t]
  \centering
    \includegraphics[width=1.\linewidth, trim=0 0 0 0, clip]{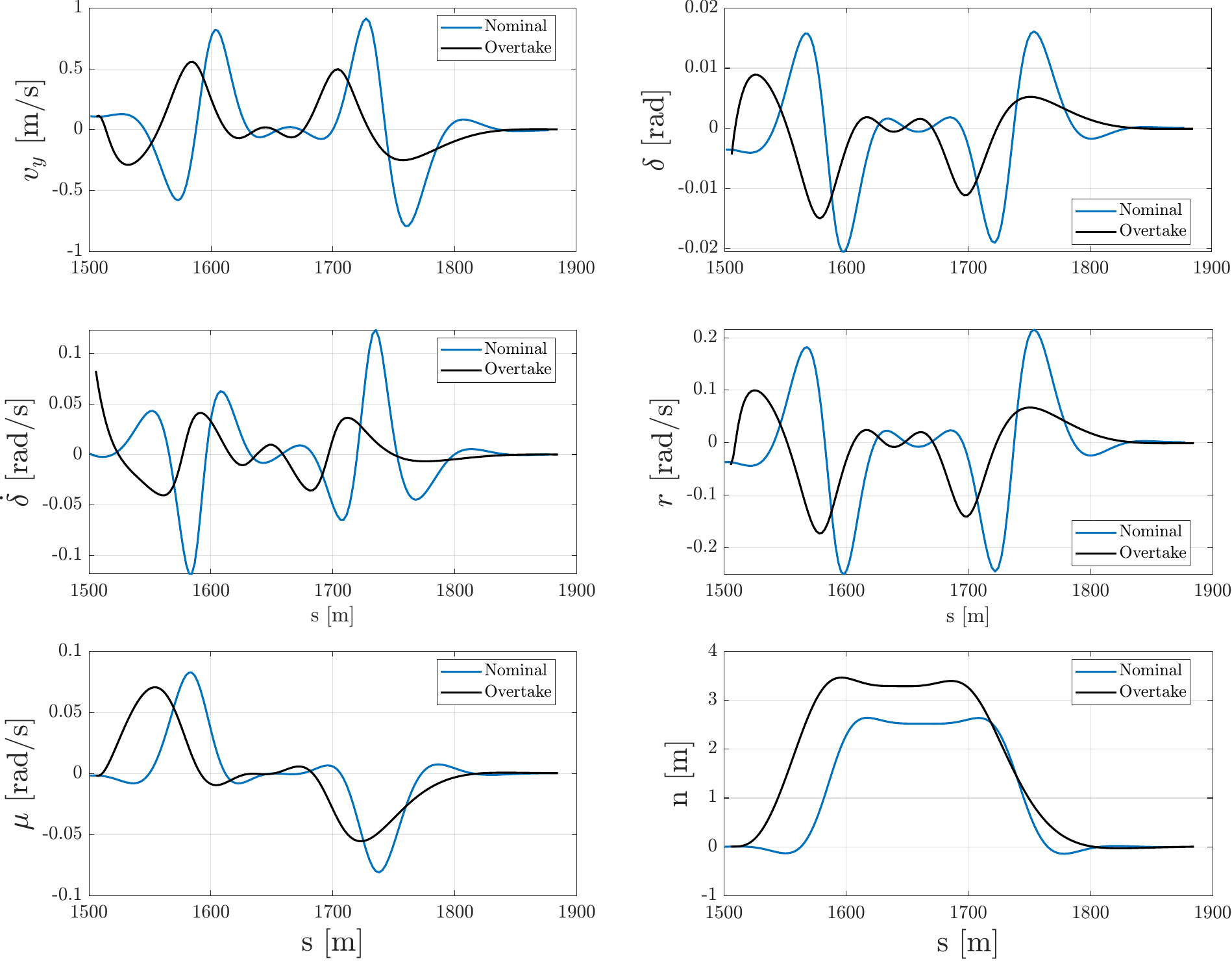}
\caption{The blue and black lines show the planner prediction horizon before and after cost switching, with weights fully shifted to the overtake task in the latter case.}
  \label{fig:cost_comparison_planner}
\end{figure}

Figure~\ref{fig:cost_comparison_planner} compares the planner predictions for an overtaking maneuver in the simulation environment. For generality, an avoidance maneuver on a straight section is considered.

Blue lines indicate the maneuver with nominal weights, where path tracking and aggressiveness are prioritized for single-vehicle lap-time performance. In contrast, the overtaking weights promote smoother control actions, as reflected by the predicted steering ($\delta$) and its derivative ($\dot{\delta}$), whose peak value decreases from 0.11 to 0.036 $\frac{rad}{s}$. At the same time, they allow larger deviations from the global path, as shown by the predicted lateral error ($n$), which increases from 2.64 to 3.46 m.
As a result, the maneuver becomes less aggressive, with lower values of $v_y$ and $r$, while still allowing a safe overtaking action. For instance, the peak lateral velocity decreases from 0.92 to 0.49
$\frac{m}{s}$.

\begin{figure}[t]
  \centering
  \begin{minipage}{1.\columnwidth} 
      \centering
      \includegraphics[width=1.\columnwidth, trim=0 0 0 0, clip]{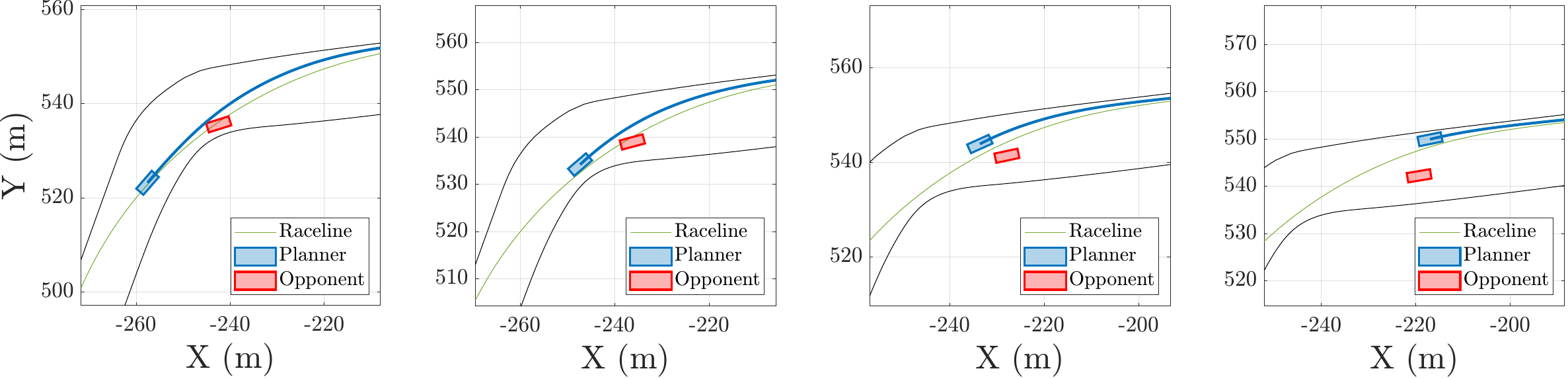}
      \subcaption{Overtake on Turn 8.}\label{fig:contructor}
  \end{minipage}

  \vspace{0.4cm} 

  \begin{minipage}{1.\columnwidth}
      \centering
      \includegraphics[width=1.\columnwidth, trim=0 0 0 0, clip]{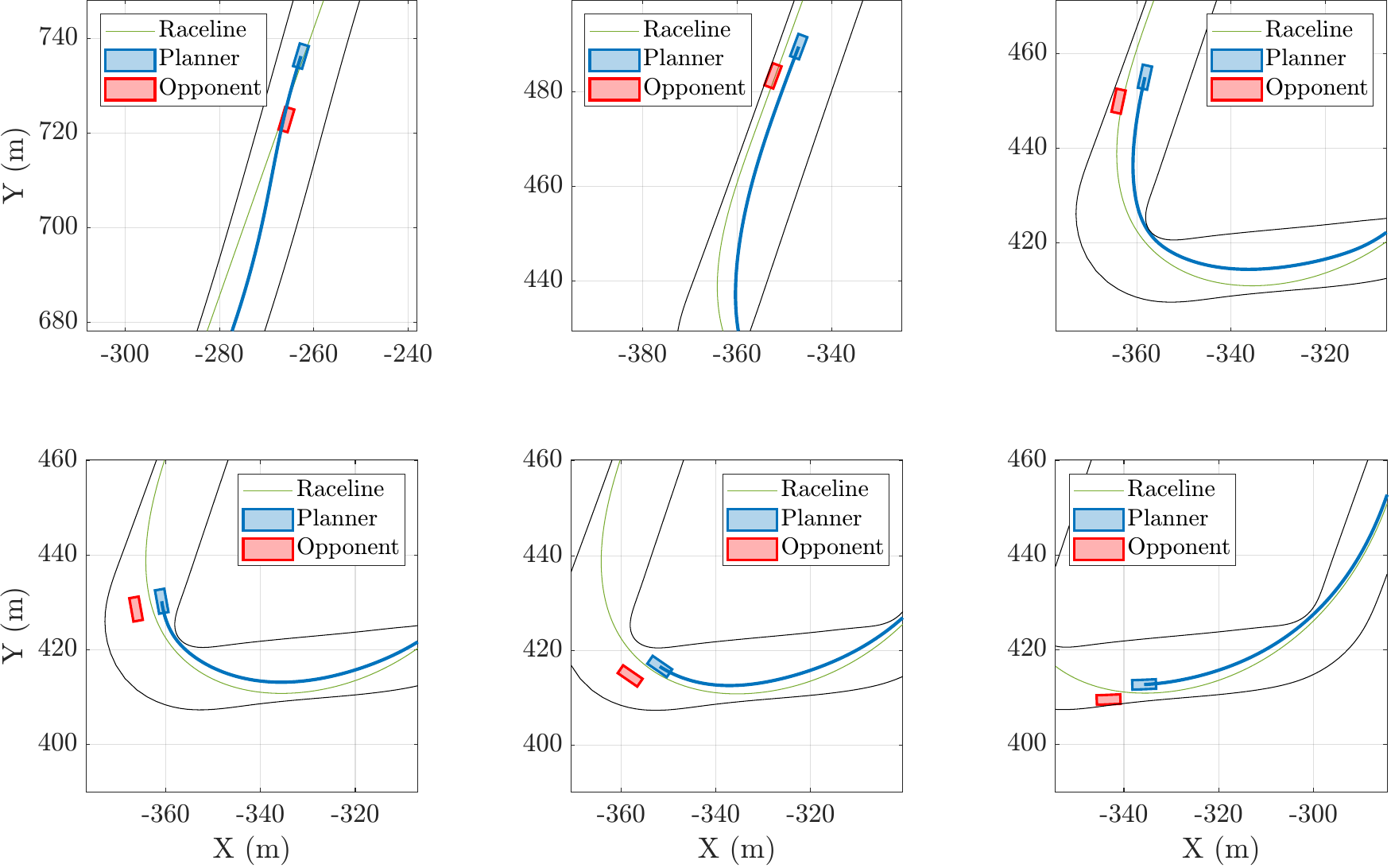}
      \subcaption{Overtake on Turn 6.}
      \label{fig:tum}
  \end{minipage}
\caption{Overtake maneuvers in a real-world context.}
  \label{fig:overtake_real}
\end{figure}

The proposed solution performed smooth yet effective overtaking maneuvers within real scenarios as well.
In particular, Figure~\ref{fig:overtake_real} shows two overtaking maneuvers during the Gran Finale event of A2RL. The blue line indicates the planned trajectory, while the red box represents the opponent’s position. The green line denotes the global racing line.

\subsection{Oversteer Scenario}
\label{subsec:res_oversteer}

\begin{figure}[ht!]
  \centering
    \includegraphics[width=.97\linewidth, trim=0 0 0 0, clip]{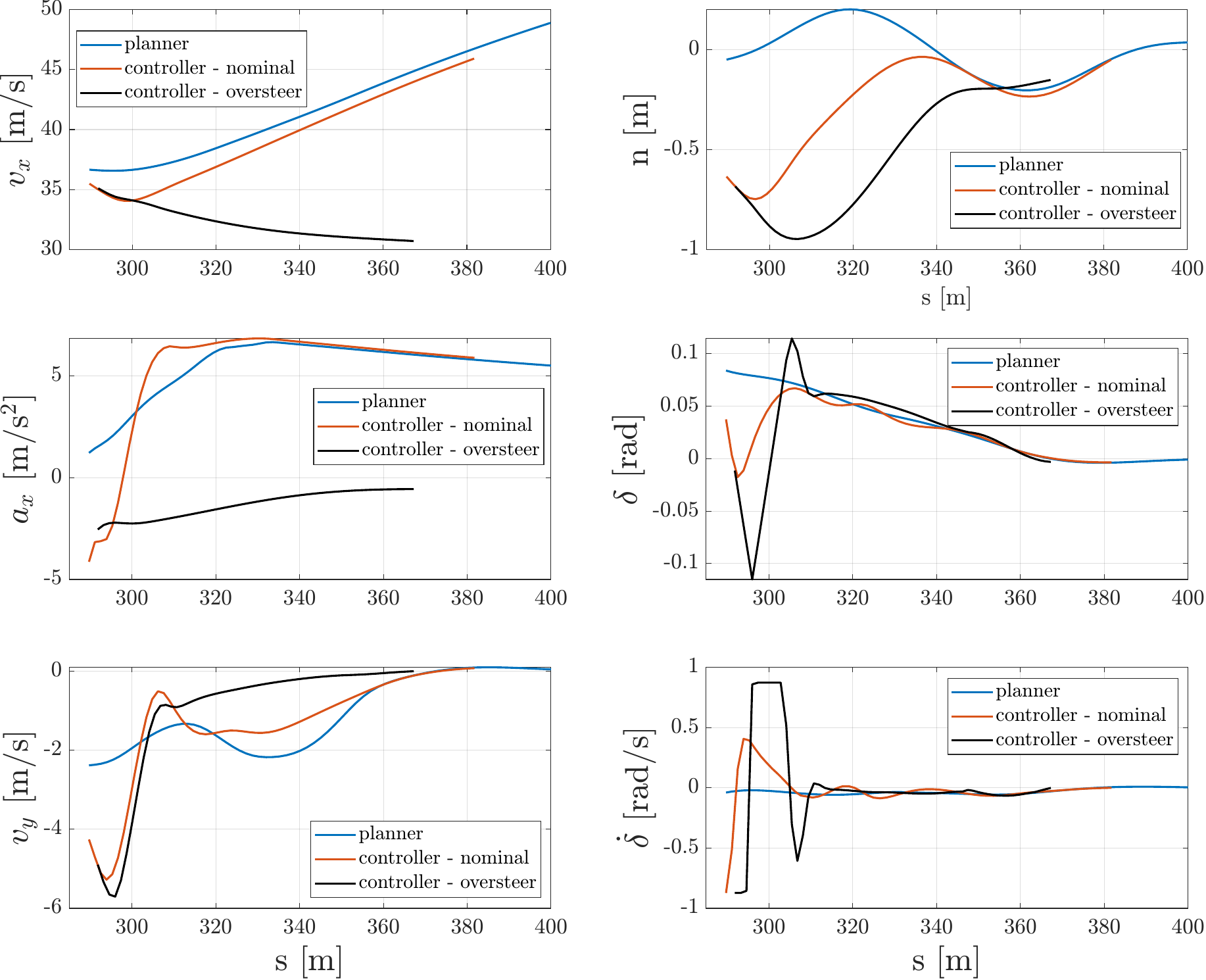}
\caption{The blue line denotes the planner prediction horizon (reference), while the orange and black lines show the controller prediction horizon before and after cost switching, with weights shifted to the oversteer task in the latter case.}
  \label{fig:cost_comparison}
\end{figure}

\begin{figure}[h!]
  \centering
  \begin{minipage}{1.\columnwidth} 
      \centering
      \includegraphics[width=1.\columnwidth, trim=0 0 0 20, clip]{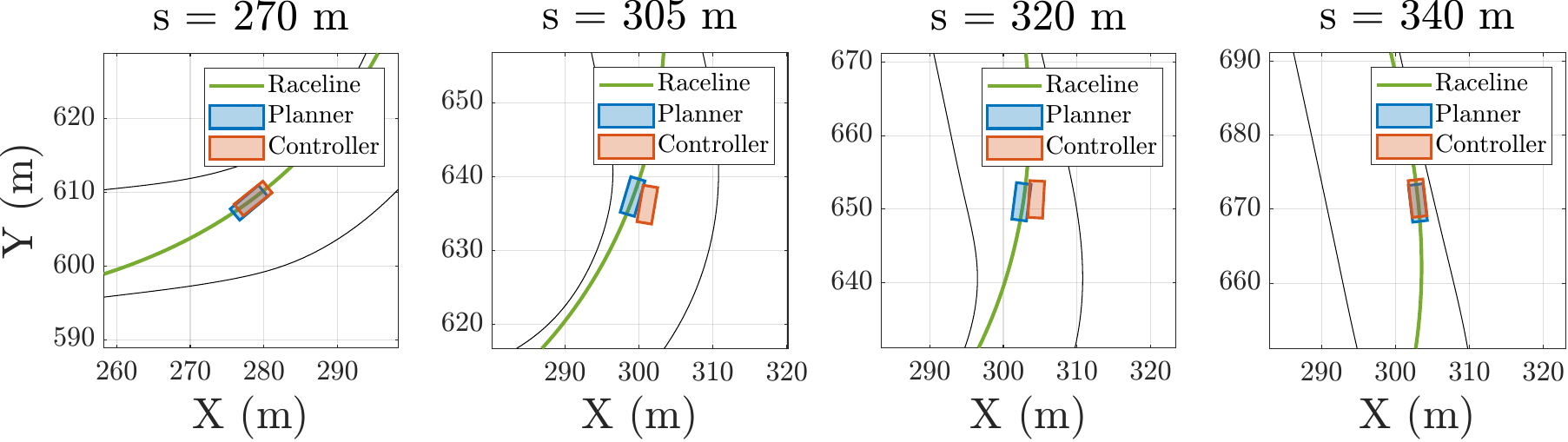}
      \subcaption{Planned and controller pose.}\label{fig:t1_oversteer_sub1}
  \end{minipage}

  \vspace{0.4cm} 

  \begin{minipage}{1.\columnwidth}
      \centering
      \includegraphics[width=1.\columnwidth, trim=0 0 0 0, clip]{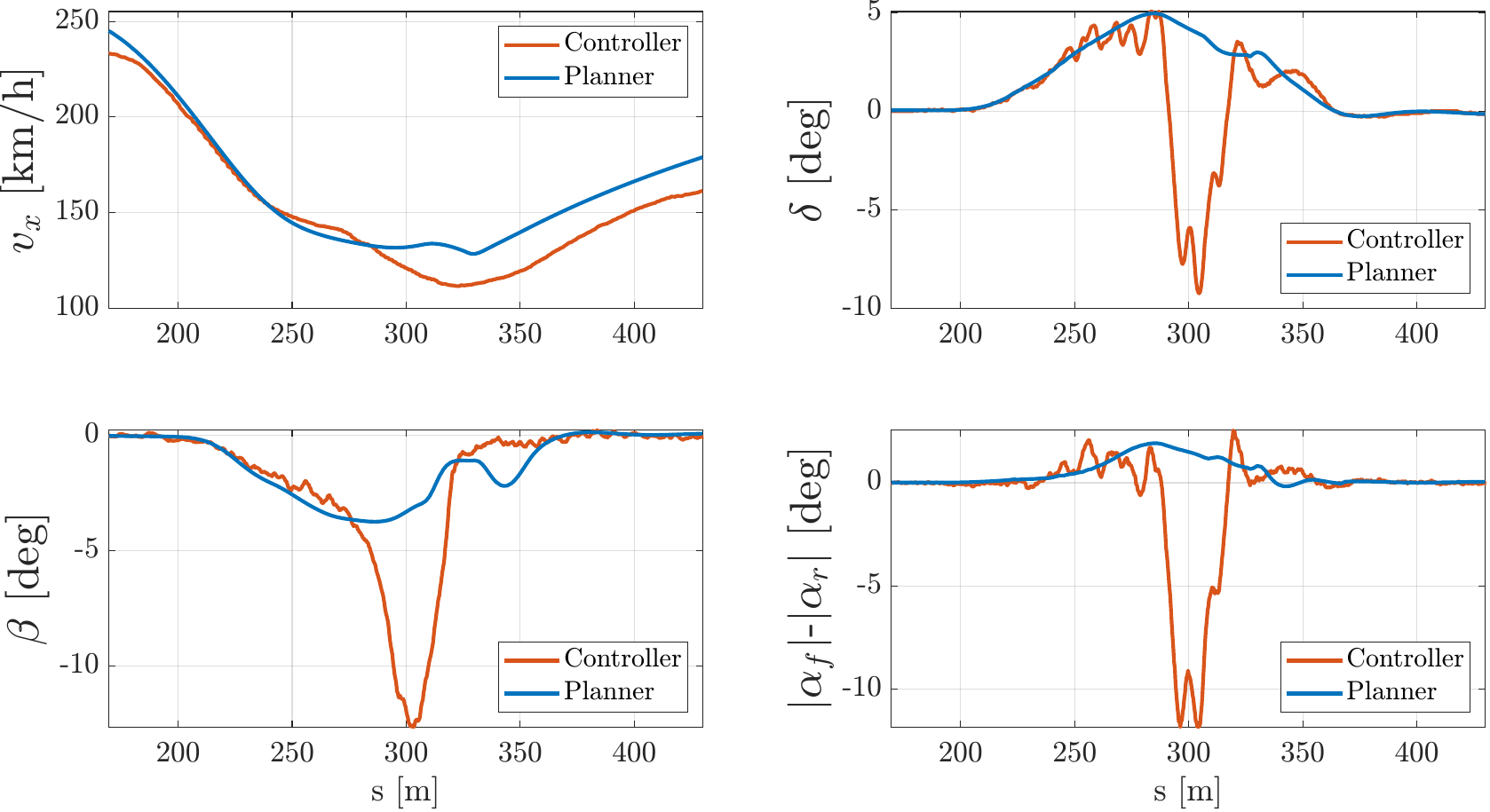}
      \subcaption{Planned and controller longitudinal velocity ($v_x$), steering ($\delta$),
      sideslip angle ($\beta$) and understeer angle ($|\alpha_f|$-$|\alpha_r|$).}
      \label{fig:t1_oversteer_sub2}
  \end{minipage}
\caption{Vehicle state differences between planner prediction and closed-loop controller response.}
  \label{fig:t1_oversteer}
\end{figure}

Cost-function blending proves crucial also for stability recovery, as highlighted in Figure~\ref{fig:cost_comparison}. During the oversteer event, the vehicle drifts outward, generating a negative lateral error while the reference remains on the left. With the nominal cost function (orange), velocity and lateral tracking, together with steering smoothness, dominate the optimization, resulting in a potentially insufficient countersteer ($\delta$). 
The black line represents the controller horizon in the subsequent planning iteration (0.04 s later), with the updated cost function. Here, the velocity and lateral error terms are temporarily down-weighted, allowing larger tracking errors, with the predicted lateral displacement increasing from 0.74 m to 0.94 m, while prioritizing vehicle stabilization. The same applies to the steering rate ($\dot{\delta}$) term, enabling a faster and broader countersteer action that saturates the steering rate at the imposed MPC limits. As a result, the steering command focuses on stability recovery rather than strict path tracking, with the planned steering angle increasing in magnitude from -0.017 rad to -0.11 rad.

Figure~\ref{fig:t1_oversteer} shows a real-world oversteer event in Turn~1.
The upper subplot (\ref{fig:t1_oversteer_sub1}) show the difference between the planned (blue) and 
actual (red) position and heading.
The planner stays consistent with the global racing-line (green), beside unnoticeable adjustment, while the controller exhibits a high heading angle (subplot 2) compared to the planned.
In the lower subplot (\ref{fig:t1_oversteer_sub2}), the planned trajectory serves as the baseline reference. In this scenario, the vehicle enters the corner at a slightly higher speed than the local planner calculates (s $\approx 250$ m), causing instability. 
Consequently, the understeer angle briefly inverts the trend, becoming highly negative around $s \approx 300$~m. The controller promptly countersteers, limiting the sideslip angle to approximately 
$\beta \approx 13^\circ$.
Stability is recovered by $s \approx 320$~m, with minimal oscillations thereafter.

\section{CONCLUSIONS}
\label{sec:conclusions}
The presented solution, focusing on detecting and switching between different driving contexts, exploits the strengths of MPC-based solutions, solving the inherent problems related to the path tracking-oriented cost function used in autonomous racing.
Furthermore, a dual-MPC approach is used, maximizing the strengths of open-loop versus closed-loop control, and logical task assignment to each MPC.
The solution can be adapted to new driving contexts, such as urban roads and highways, focusing on additional scenarios like comfort and jerk minimization. Other driving styles could be needed for edge cases in racing, such as mechanical malfunctions (e.g., brakes or steering failure) or performance optimization on particular track sectors (e.g., high decelerations or extremely tight corners). Future work will also focus on automating the weights tuning and the cost switch decisions, as well as extending the overall architecture presentation and results. 







\bibliographystyle{unsrt}   
\bibliography{references}   

\end{document}